\documentclass{article}

\newif\ifzhdraft
\zhdraftfalse         % English version

\usepackage{iclr2027_conference}

\usepackage{times}

\ifzhdraft
  \usepackage{xeCJK}
  \setCJKsansfont{FandolHei-Regular.otf}
  \setCJKmonofont{FandolFang-Regular.otf}
\fi

\usepackage{amsmath,amsfonts,bm}

\def\eqref#1{equation~\ref{#1}}
\def\1{\bm{1}}

\DeclareMathAlphabet{\mathsfit}{\encodingdefault}{\sfdefault}{m}{sl}
\SetMathAlphabet{\mathsfit}{bold}{\encodingdefault}{\sfdefault}{bx}{n}

\usepackage{amsmath,amssymb,amsthm}
\usepackage{graphicx}
\usepackage{booktabs}      % 三线表
\usepackage{flafter}
\usepackage{placeins}
\usepackage{multirow}
\usepackage{makecell}
\usepackage{xcolor}
\usepackage{hyperref}
\hypersetup{hidelinks}  % no coloured boxes around links
\usepackage{url}
\usepackage{arydshln}
\usepackage{xspace}
\newcommand{\method}{CARE-PO\xspace}
\newcommand{\methodlow}{\method (low)\xspace}
\newcommand{\methodhigh}{\method (high)\xspace}

\newcommand{\gup}[1]{\textcolor{green!55!black}{\scriptsize$\blacktriangle$#1}}
\newcommand{\gdn}[1]{\textcolor{green!55!black}{\scriptsize$\blacktriangledown$#1}}
\newcommand{\rup}[1]{\textcolor{red!75!black}{\scriptsize$\blacktriangle$#1}}
\newcommand{\rdn}[1]{\textcolor{red!75!black}{\scriptsize$\blacktriangledown$#1}}

\newcommand{\todoNote}[1]{}

\title{CARE-PO: Controllable Tool-Call Efficiency for LLM Agents}

\author{%
\textbf{Xiaoyang Ye}$^{1,2}$, \textbf{Xiao Yao}$^{1,2}$, \textbf{Hairui Zhu}$^{1}$, \textbf{Yiying Yang}$^{1}$, \textbf{Bei Gan}$^{3}$,\\
\textbf{Jinliang Lu}$^{3}$, \textbf{Tengjin Weng}$^{1,2}$, \textbf{Yuan Yao}$^{1}$, \textbf{Wenhao Jiang}$^{1}$\\[6pt]
$^{1}$Guangdong Laboratory of Artificial Intelligence and Digital Economy (SZ)\\
$^{2}$Shenzhen University\\
$^{3}$China Southern Power Grid Company Limited
}

\iclrfinalcopy

\newcommand{\groupdash}{\noalign{\vskip 1pt}\hdashline\noalign{\vskip 1.5pt}}

\begin{document}
\maketitle
\lhead{Under review as a conference paper at ICLR 2027}

\begin{abstract}
External tool calls are a major running cost of an LLM agent, while its training objectives align almost exclusively with task completion. In a step-level diagnosis of Qwen3.5 trajectories at two model sizes on GAIA, a long-horizon tool-use benchmark, we find that most contain removable calls, at a similar rate at both sizes. Some methods add an efficiency reward to the objective to reduce the call count, but introduce problems of their own: the anchor they price calls against is computed from the policy's own rollouts (e.g., the group's shortest correct rollout) and so moves with the policy; training can drift toward a collapsed policy that issues no tool call at all, a degeneracy seen in published reports and in our 27B baseline; and a training run delivers a single operating point, so the user cannot trade efficiency against capability.
We present \textbf{CARE-PO} (Controllable Anchored Relative Efficiency Policy Optimization) to mitigate these problems. The efficiency reward prices calls against a reference budget anchored to the task's own difficulty and fixed before training.
Injecting a successful trajectory from the training data into each online GRPO rollout group and charging for evidence
a trajectory never gathers ease the drift toward a collapsed call count. Shaping the cost curve by an effort tag in the prompt
yields an efficiency and a capability operating point from a single training run.
On GAIA, five of the six operating points across three model sizes improve on their base model in success rate and call
count at once. On four held-out multi-hop benchmarks the result is size-dependent: at 27B
and 4B success rate is held or raised at a lower call count, while at 9B both operating points trade success rate
for fewer calls. The call count remains well above zero throughout training at all three sizes.

\end{abstract}

\section{Introduction}
\label{sec:intro}

In the ReAct-style loop of an LLM agent~\citep{yao2023react}, the cost of a tool call \emph{compounds}: its observation stays in the context and is re-processed on every later turn, so one low-yield call adds cost and latency to every remaining turn.

\begin{figure}[!tb]
\centering
\includegraphics[width=\textwidth]{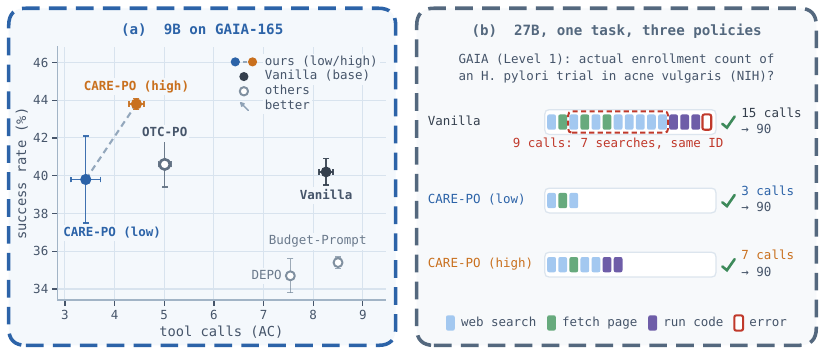}
\caption{(a) Success rate against average tool calls (AC) at 9B on GAIA's 165-question validation split; error bars
are the s.d.\ over three runs. Vanilla is the untrained model; CARE-PO (low) and (high) are two operating points of one
set of weights; the baselines are defined in Section~\ref{sec:setup}. (b) On one GAIA task at 27B, all three answer
correctly, with 15, 3 and 7 calls (Appendix~\ref{app:case}).}
\label{fig:teaser}
\end{figure}

We observe that low-yield calls are pervasive at both model sizes we examine. In a step-level diagnosis of 221 GAIA
trajectories~\citep{mialon2023gaia} from Qwen3.5-9B and 27B, 82.8\% contain a removable
(deletable or mergeable) call; those trajectories average 5.7 removable calls, and 68.0\% of all calls are removable
(Section~\ref{sec:diagnosis}).
The rate is similar at both sizes, pointing less to any one model's competence than to the
training objective: tool-use training targets when to call and how to use the result, not what a
call costs~\citep{schick2023toolformer,qin2024toolllm}; RL rewarding only final correctness makes extra exploration rational~\citep{jin2025searchr1,wang2025otc}; and agentic post-training for long-horizon
tasks encourages continued calling, with decorative calls and tool overuse
documented~\citep{ma2025proofofuse}, compounding an over-calling bias that precedes agentic training~\citep{shi2026overcalling,wu2026tocall}.

Existing interventions on call cost~\citep{bats2025,chen2025depo,wang2025otc,li2026tepo} either leave the budget to the user as an external constraint or fix the efficiency--capability trade-off at training time, \emph{one set of weights per
operating point}. A separate line of work lets the user choose the model's effort at
inference~\citep{l12025,thinkdial2025}, but this choice mostly concerns reasoning tokens; concurrent
extensions of that choice to tool calls leave the training objective
unchanged~\citep{chen2026steering,fang2026budget}.
A reasonable call count is set by two factors: how many calls the task requires, and how many calls this deployment is willing to spend for success rate. The first is independent of the policy and should not be estimated from the
policy's own samples; the second is a deployment choice and should not be fixed in training.

We therefore treat call efficiency as an \emph{inference-time controllable property} and propose
\textbf{CARE-PO}, with three contributions:
\textbf{(1) A compression yardstick anchored to each task's difficulty.} Each task's \emph{reference budget}, set by an offline pass before training,
anchors the efficiency term. The cost curve around it is bounded and monotone, so a misestimated budget changes the magnitude of the pressure, not its direction, and compression below the budget earns further credit; calls saved
without obtaining evidence earn nothing (Sections~\ref{sec:method:bref} and~\ref{sec:method:reward}).
\textbf{(2) Call-count collapse under efficiency rewards is effectively eased.}
Each of the two routes to a policy that issues no call is addressed by one design: on the reward side, the evidence
pricing just described; on the optimizer side, where a group uniform in calls leaves the advantage nothing to
compare~\citep{avspo2026}, each group receives the task's successful trajectory from the training data, its \emph{source trajectory}, as a floor (Section~\ref{sec:method:anchor}). At 27B an
online-RL baseline trained with a conventional efficiency reward collapses toward issuing no call partway through training; ours does not at any size, with the same learning rate and number of training steps (Section~\ref{sec:analysis:collapse}).
\textbf{(3) The efficiency--capability trade-off is decided at deployment.}
The reward prices a call differently at each effort tier, so one set of weights learns an efficiency and
a capability operating point, selected at inference by a low or high effort tag in the prompt (Section~\ref{sec:method:effort}).

On GAIA, five of the six operating points across three model sizes improve on their base model in success rate and
call count at once (Table~\ref{tab:main}, Figure~\ref{fig:teaser}); the sixth matches it in success rate at 58.6\% fewer calls. On four held-out multi-hop benchmarks
the efficiency point calls roughly a quarter to a half as often; success rate is held or raised at 27B and 4B, while at 9B both points trade success rate for fewer calls.

\section{Related Work}
\label{sec:related}

\textbf{Reducing the number of tool calls.} Overthinking and early-exit control of reasoning tokens are well
studied~\citep{cuadron2025overthinking,wei2025rcp}; external tool calls, much less so. Tool overuse is
documented~\citep{qian2025smart}, but mostly as trajectory-level counts, rarely with a step-level
criterion for locating redundancy inside a trajectory, which Section~\ref{sec:diagnosis} provides. Existing interventions divide into three classes by where they act, increasingly close to ours. \emph{Inference-time} ones leave the weights untouched and impose the budget as an external constraint~\citep{bats2025,chen2026steering,fang2026budget}; \emph{offline training} changes the weights on a pre-collected signal that cannot adapt to online rollouts~\citep{chen2025depo,webclipper2026}; \emph{online RL} optimizes with GRPO~\citep{shao2024deepseekmath} and places the efficiency demand entirely on the reward, as in OTC-PO~\citep{wang2025otc} and TEPO~\citep{li2026tepo}.
This class measures efficiency against the policy's own samples: OTC-PO's anchor, the group's minimum call count among correct trajectories, drifts with the policy, while TEPO measures efficiency by the fraction of entropy-reducing calls and sets no task-level reference. Our cost is instead measured against a per-task reference budget, fixed before training (Section~\ref{sec:method:bref}). Both methods also yield a single operating point, so another requires retraining.
An efficiency reward can also equalize a rollout group's scores, leaving GRPO no gradient~\citep{avspo2026}; search
agents have been reported to collapse in training~\citep{lld2025}, and per-call rewards have been proposed to counter
policies that reduce calls indiscriminately~\citep{ma2026trm}.

\textbf{Selecting effort at inference.} Controlling a model's reasoning effort is an emerging trend in
pure reasoning settings~\citep{l12025,fang2025thinkless,zhang2025adaptthink}: a discrete mode in the system
prompt lets the user select among a few operating points of one set of weights~\citep{thinkdial2025}, adjusting
internal reasoning tokens rather than external tool calls.
The inference-time interventions above extend that choice to tool
calls~\citep{chen2026steering,fang2026budget} but leave the training objective unchanged, so no tier is trained.
We instead write the tier into the objective, so the two operating points are learned rather than
imposed at inference (Section~\ref{sec:method:effort}).

\section{Method: \method}
\label{sec:method}

\begin{figure}[!tb]
\centering
\includegraphics[width=\textwidth]{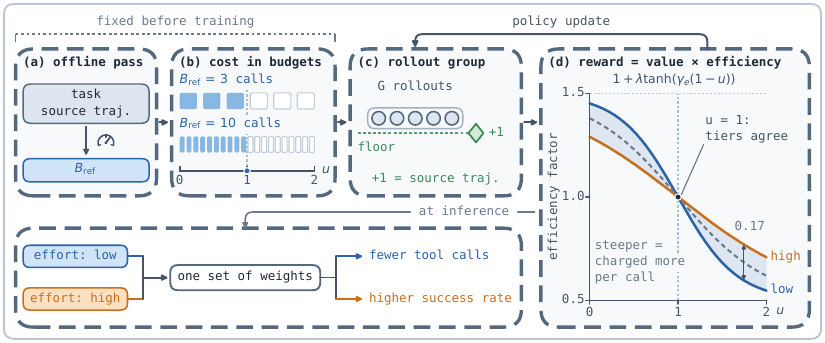}
\caption{\method overview. The budgets 3 and 10 in (b) are illustrative. Cost $u$ covers more than calls:
Eq.~(\ref{eq:u}) also charges a repeated call twice and charges evidence the trajectory never obtained.
The 0.17 marked in (d) is the largest gap between the tiers, and the dashed middle curve there is the
unconditioned tier (no tag). Definitions in Sections~\ref{sec:method:problem}--\ref{sec:method:anchor}.}
\label{fig:overview}
\end{figure}

This section first diagnoses how redundant calls make a trajectory costly, which decides where the reward should act (Section~\ref{sec:diagnosis}). Sections~\ref{sec:method:problem} and~\ref{sec:method:bref} then state the objective and the reference budget that captures a task's difficulty, Section~\ref{sec:method:reward} turns the objective into a reward and sets the price of a call at each effort tier, and Section~\ref{sec:method:anchor} adds the floor that eases call-count collapse; Figure~\ref{fig:overview} gives an overview.

\subsection{Diagnosis of Redundant Calls}
\label{sec:diagnosis}

Of 221 GAIA trajectories collected for this diagnosis from Qwen3.5-9B and 27B, 183 (82.8\%) contain a call
deletable outright or mergeable into its neighbour---85.7\% at 9B, 80.2\% at 27B; those trajectories average 5.7
removable calls, and 68.0\% of all calls are removable without breaking the answer's evidence
chain (procedure and judge in Appendix~\ref{app:diagnosis}). None of the 221 enters training.

These rates measure the amount of redundancy, not its cost to a trajectory, and it is the cost that decides
where the reward should act. Step-by-step labelling yields many behavioural patterns; grouping
them by \emph{what each costs the trajectory} leaves only three forms.

\emph{(i) Calls with no increment} return no task-relevant new information: aimless search,
re-verification of a confirmed result, one acquisition split into several. Each adds a call and no
information, so the call count measures them fully.
\emph{(ii) Repeated calls} closely match an earlier call in tool, arguments and return, re-consuming an observation
already obtained; the count cannot distinguish them from productive ones. Repetition occupying a trajectory's tail is a \emph{tail repetition loop}: it often persists to the call cap~\citep{liu2024agentbench},
which wastes calls and also leaves the trajectory unable to answer.
\emph{(iii) Evidence gaps} belong to no single call: a cue that could advance the task appears and goes unused.
The side searches that follow are counted as form (i), but the harm is to correctness rather than to cost. Frequency and harm run in opposite orders across the three: form (i), fully priced by the count, is the most common and least harmful, while forms (ii) and (iii), whose harm the count misses, are most associated with wrong answers (definitions, criteria and per-form statistics in Appendix~\ref{app:diagnosis}). Section~\ref{sec:method:reward} therefore prices each of these two with a term of its own.

\subsection{Problem Setting and Objective}
\label{sec:method:problem}

In the ReAct-style loop~\citep{yao2023react}, the policy $\pi_\theta$ alternates reasoning and tool calls on a task
$x$ until it answers. A trajectory $o$ makes $S(o)$ tool calls under a hard cap $T_{\max}$;
$c(x,o)\in\{0,1\}$ marks whether $o$ \emph{delivers}, i.e.\ completes the task goal (graded automatically,
Appendix~\ref{app:impl}).

The goal is to reduce $S(o)$, and no trajectory may trade correctness for efficiency. Policy-gradient training optimizes an
unconstrained expected return, so that requirement enters as a product: return is \emph{delivered value} times an
\emph{efficiency factor}, and a trajectory that fails to deliver scores below every delivering one however many
calls it saves; an additive step penalty does not
guarantee this. The cost the efficiency factor measures cannot be the raw call count, because tasks differ widely in how many calls they need.
We measure each training task's difficulty by its \textbf{reference budget} $B_{\mathrm{ref}}(x)$, an estimate of the
calls needed to complete it, and define cost as $u=S(o)/B_{\mathrm{ref}}(x)$, the number of budgets a trajectory
spends (Eq.~(\ref{eq:u}) also charges repeated calls and evidence never obtained). With a budget of 3 one call adds $1/3$ to $u$, with a budget of
10 only $1/10$ (Figure~\ref{fig:overview}(b)), so pressure decreases as the task demands more. The estimate is made
before training begins (Section~\ref{sec:method:bref}) and then stays fixed, so the demand on the call count never
tightens as the policy's trajectories shorten.

Once cost is defined, two choices remain free, how strongly to penalize overspending and how much to reward saving, and the effort tier sets them. The objective is therefore a
family, indexed by an \textbf{effort tier} $e\in\mathcal{E}=\{\varnothing,\text{low},\text{high}\}$ that is an input to the policy: $\varnothing$ is the unconditioned tier, with no effort tag in the prompt; low and high are the two tiers the user selects at inference.
Formally,
\begin{equation}
\label{eq:objective}
\pi_e^{\star}
=\arg\max_{\pi}\
\mathbb{E}_{x\sim\mathcal{D},\ o\sim\pi(\cdot\mid x,\,e)}
\Big[\,
V_e(x,o)\cdot U_e(u)
\,\Big],
\end{equation}
where $V_e$ is the delivered value and $U_e$ the efficiency factor, monotonically decreasing in the cost $u$;
Section~\ref{sec:method:reward} gives their concrete forms and what $e$ changes in them. One conditional policy
$\pi_\theta(\cdot\mid x,e)$ approximates the whole family, so the two operating points share one set of weights.

\subsection{Training Data and Reference Budget}
\label{sec:method:bref}

\textbf{Training data.} Training tasks come from MiroVerse~\citep{miroverse2025} (v0.1, about 148k trajectories):
long, call-dense and often redundant, yet automatically gradable. Most of it is Voyager1.0, the rest mainly
multi-hop QA and web-browsing datasets such as MuSiQue, HotpotQA, 2WikiMultihopQA and WebWalkerQA. Preprocessing,
which never involves the policy, runs in three steps. Successful source trajectories with at least five calls form
the candidate pool (about 64k), since shorter ones leave almost nothing to compress; the offline pass below gives each
candidate a reference budget, an evidence checklist and a gradable answer; and three checks on these quantities ensure
that coverage is not binary, that each piece of evidence is worth a call and that the source trajectory leaves room to
compress (Appendix~\ref{app:impl}). Training draws its tasks from a fixed list of 300, balanced across reference budgets
and shared by all three sizes. Exact and 8-gram matching finds no training question in any evaluation set.

\textbf{Reference budget and evidence checklist.} One offline pass per task estimates, then verifies, the reference
budget and the evidence checklist, and discards tasks with no extractable gradable answer. Guided by the three forms of Section~\ref{sec:diagnosis}, a judge model reduces the
source trajectory to a minimal chain of essential steps and marks the evidence each requires: the chain's length is
the candidate value, and the marked entities drawn from tool observations form the \textbf{evidence checklist} $E(x)$.
A second prompt, which sees the chain but not how it was estimated, checks it: a failure triggers re-estimation
under a relaxed budget, and a task whose budget must be relaxed to the source trajectory's call count
$S_{\mathrm{long}}(x)$ is discarded as having no compressible slack (details in Appendix~\ref{app:impl}). The pass outputs $B_{\mathrm{ref}}(x)$, $E(x)$ and a gradable answer for
training; the chain itself is never shown to the model. The reference budget need only be reachable in real
execution, not the task's true minimum or any one model's optimum: an estimation error changes the magnitude of the pressure, not its direction.

\subsection{\method Reward: Delivered Value Times Task-Relative Efficiency}
\label{sec:method:reward}
\label{sec:method:effort}

\method keeps the GRPO optimizer; each trajectory's reward is the product of the factors of
Eq.~(\ref{eq:objective}):
\begin{equation}
\label{eq:reward}
R_e(x,o)=\underbrace{\Big[c(x,o)+\big(1-c(x,o)\big)\,b_e\,\mathrm{Cov}(o)\Big]}_{\text{delivered value}\ V_e}\cdot
\underbrace{\Big[1+\lambda\tanh\big(\gamma_e(1-u)\big)\Big]}_{\text{efficiency factor}\ U_e} .
\end{equation}

\textbf{Delivered value.} A delivering trajectory is worth 1; a failing one earns partial credit $b_e\,\mathrm{Cov}(o)$,
where $\mathrm{Cov}(o)\in[0,1]$ is the fraction of the evidence checklist $E(x)$ its observations cover
(Appendix~\ref{app:impl}). A trajectory that terminates early gathers little evidence and earns little, so no separate early-termination penalty is needed---a
call-free trajectory obtains no evidence and is worth exactly zero unless correct. Partial
credit must never lift a failure above a success: a failing trajectory earns at most $b_e(1+\lambda)$ (full coverage,
efficiency factor at its upper bound) and a delivering one at least $1-\lambda$, so $b_e<(1-\lambda)/(1+\lambda)$
suffices to keep that ordering for every task, tier and call count. We set $\lambda=0.5$, giving $b_e<1/3$.

\textbf{Task-relative efficiency.} Three properties of the tanh's shape govern the pressure toward shorter
trajectories. It is strictly decreasing in the cost $u$, so compression below the reference budget earns further
credit. It is bounded on both sides, confining the factor to $1\pm\lambda$, so neither the penalty for waste nor the
return on extreme compression grows without bound. Finally, it is steepest at $u=1$, where calls exactly meet the reference budget, placing the pressure where the choice between one more call and answering now is hardest.
The cost is defined so that calls saved without obtaining evidence earn nothing:
\begin{equation}
\label{eq:u}
u=\frac{S_{\mathrm{charged}}(o)}{B_{\mathrm{ref}}(x)}+\big(1-\mathrm{Cov}(o)\big),
\qquad
S_{\mathrm{charged}}(o)=S(o)+\big|\{j:\ \text{call } j \text{ repeats an earlier call}\}\big| .
\end{equation}
The first term prices form (ii) of Section~\ref{sec:diagnosis} by charging repeated calls twice. The second term,
the \emph{coverage term}, prices form (iii): evidence the checklist requires but the trajectory never obtained is
converted into reference budgets in proportion, so missing half the evidence counts as spending half a budget more.
A trajectory that calls no tool thus costs exactly one reference budget even when it happens to be correct---the
same as one that spent its entire budget delivering.

\textbf{Tier control: changing the price of a call.} Task difficulty and effort tier meet in one quantity: near $u=1$, one more call lowers the
efficiency factor by about $\lambda\gamma_e/B_{\mathrm{ref}}(x)$, so a call is cheaper on a harder task and its price
scales with the tier's $\gamma_e$. A single \emph{price factor} $t>1$ sets
$\gamma_e$: $t$ at the low tier, $1$ at the unconditioned tier and $1/t$ at high, with $t=1.5$, so a call at the low
tier costs $t^2$ times what it costs at the high tier. Partial credit scales inversely, $b_e\propto 1/\gamma_e$: we set $b_{\text{high}}=0.30$, giving $b_\varnothing=0.20$ and $b_{\text{low}}=0.13$, all three within the bound
$b_e<1/3$ above. The two effects work together: the
high tier credits partial progress more and charges each call less. Only the low and high tiers are
exposed to the user, a binary mode with precedent in reasoning-mode
switches~\citep{fang2025thinkless,zhang2025adaptthink,qwen3}.

\textbf{The tier-neutral point.} At $u=1$ the argument of the tanh is zero, so the efficiency factor is 1 for any
$\gamma_e$: a delivering trajectory that spends exactly one reference budget scores the same at every tier. A larger $\gamma_e$ makes the curve steeper, which fixes
the tiers' ordering: below one reference budget the low tier scores highest, above it lowest, the separation widening
away from the neutral point and narrowing again once overspending saturates the curves (Figure~\ref{fig:overview}(d)).
For failing trajectories, the value $b_e\,\mathrm{Cov}(o)$ is higher at the high tier than at the
low one, but only through scaling the partial credit, independently of the call count.

\textbf{Conditioning and mixing.} The effort tag sits in the prompt, and the difference between the two operating points rests entirely on it. The effort tier is sampled independently for each rollout,
with probabilities $0.2:0.4:0.4$ for the unconditioned tier, low and high, so a group mixes tiers and
the group-relative advantage compares them within a task, against the common reference of the tier-neutral point.
Mixing tiers also widens the spread of scores within a group, on which the group-relative advantage depends (Section~\ref{sec:method:anchor}). The unconditioned tier sets the default behaviour with no tag and serves as an
internal control (Section~\ref{sec:analysis:effort}).

\subsection{The Source Trajectory as a Group Floor}
\label{sec:method:anchor}

Pricing tool calls admits a degenerate solution: issuing no call at all. The reward side is already addressed by
evidence pricing, the coverage term of Eq.~(\ref{eq:u}); the optimizer side stays open. The group-relative advantage
scores a rollout against the rest of its group, so once the whole group issues the same number of calls, the price
of a call no longer distinguishes rollouts~\citep{avspo2026}.

Each group of $G$ rollouts therefore receives the task's own source trajectory as member $G{+}1$, scored by
Eq.~(\ref{eq:reward}) at the unconditioned tier; injecting an extra member has precedent~\citep{avspo2026}, but this
one is fixed per task and does not move with the group. This member was not drawn from $\pi_\theta$; its importance ratio
starts at 1 and its advantage is clipped below at zero. It is a floor rather than a target. In most groups it costs more and scores below the group mean, and its clipped advantage adds no gradient; when most of the group fails and the mean drops below it, its positive advantage pulls the policy back to a delivering path, much as off-policy guidance does in mixed-policy GRPO~\citep{yan2025luffy}. When no rollout obtains evidence, the source trajectory raises the group mean, so failing call-free rollouts receive negative advantages.

\section{Experimental Setup}
\label{sec:setup}

\textbf{Models and training.} Base models are Qwen3.5~\citep{qwen35} at 4B, 9B and 27B; unless stated otherwise, analyses and ablations use 9B on GAIA. Training is
online GRPO through LoRA~\citep{hu2022lora}. One recipe serves all three sizes, with the learning rate chosen
per size. Training and evaluation share four tools: web search, page retrieval, Python execution and local file
reading. The call cap $T_{\max}=20$ exceeds the longest source trajectory of any training task.
Appendix~\ref{app:impl} gives hyperparameters and tool implementations.

\textbf{Benchmarks and protocol.} The main evaluation set is GAIA's 165-question validation
split~\citep{mialon2023gaia}, long-horizon tasks with heterogeneous tools; the \textbf{held-out evaluation} adds four
multi-hop QA benchmarks: 500 questions from each official split of 2Wiki~\citep{ho2020constructing},
HotpotQA~\citep{yang2018hotpotqa} and MuSiQue~\citep{trivedi2022musique}, and all 125 of
Bamboogle~\citep{press2023compositionality}. Training and evaluation both use live, non-stationary web search, so on GAIA we report mean and sample s.d.\ over three independent runs; on the multi-hop benchmarks
greedy decoding and query-level caching make a single run deterministic (significance tests in Appendix~\ref{app:impl}).

\textbf{Metrics.} SR is the success rate and AC the average tool calls per question (failures included). HV is the
hypervolume indicator~\citep{zitzler1999hv} on the success-rate--call-count plane: with the worst corner (zero success at the full call cap) as reference point, an operating point spans a rectangle of area $\mathrm{HV}=\mathrm{SR}\cdot(1-\mathrm{AC}/T_{\max})$, with SR as a fraction. It summarizes the trade-off (higher is better); a policy with zero success scores zero however few calls it makes. HV implies an
exchange rate set by $T_{\max}$ (at the 9B efficiency point, one saved call is worth about 2.4 SR points;
Appendix~\ref{app:impl}), so our main GAIA conclusions rest on SR and AC separately.

% Tables 1 and 2 are registered here so that they land at the top of the page where Section 5 begins.
\begin{table}[!t]
\centering
\footnotesize
\setlength{\tabcolsep}{6pt}
\caption{Main results on GAIA-165 (mean$\pm$sample s.d.\ over three independent runs; best per metric and size in bold). Below the dashed line: \methodlow, the efficiency point, and
\methodhigh, the capability point, from one set of weights. Arrows give the change from Vanilla ReAct at the same size (percentage points for SR,
relative for AC and HV); green marks an improvement.}
\label{tab:main}
\begin{tabular}{lccc}
\toprule
Method & SR$\uparrow$ & AC$\downarrow$ & HV$\uparrow$ \\
\midrule
\multicolumn{4}{l}{\textbf{Qwen3.5-9B}} \\
Vanilla ReAct & 40.2\scriptsize{$\pm$0.7} & 8.26\scriptsize{$\pm$0.14} & 0.236\scriptsize{$\pm$0.005} \\
Budget-Prompt & 35.4\scriptsize{$\pm$0.3}~\rdn{4.8} & 8.50\scriptsize{$\pm$0.05}~\rup{2.9\%} & 0.203\scriptsize{$\pm$0.001}~\rdn{14.0\%} \\
DEPO & 34.7\scriptsize{$\pm$0.9}~\rdn{5.5} & 7.54\scriptsize{$\pm$0.05}~\gdn{8.7\%} & 0.217\scriptsize{$\pm$0.005}~\rdn{8.1\%} \\
OTC-PO & 40.6\scriptsize{$\pm$1.2}~\gup{0.4} & 5.01\scriptsize{$\pm$0.13}~\gdn{39.3\%} & 0.304\scriptsize{$\pm$0.012}~\gup{28.8\%} \\
\groupdash
\methodlow & 39.8\scriptsize{$\pm$2.3}~\rdn{0.4} & \textbf{3.42}\scriptsize{$\pm$0.30}~\gdn{58.6\%} & 0.330\scriptsize{$\pm$0.015}~\gup{39.8\%} \\
\methodhigh & \textbf{43.8}\scriptsize{$\pm$0.3}~\gup{3.6} & 4.44\scriptsize{$\pm$0.15}~\gdn{46.2\%} & \textbf{0.341}\scriptsize{$\pm$0.002}~\gup{44.5\%} \\
\midrule
\multicolumn{4}{l}{\textbf{Qwen3.5-27B}} \\
Vanilla ReAct & 51.3\scriptsize{$\pm$0.3} & 10.66\scriptsize{$\pm$0.06} & 0.240\scriptsize{$\pm$0.002} \\
OTC-PO & 52.1\scriptsize{$\pm$2.4}~\gup{0.8} & 5.57\scriptsize{$\pm$0.21}~\gdn{47.7\%} & 0.376\scriptsize{$\pm$0.014}~\gup{56.7\%} \\
\groupdash
\methodlow & \textbf{57.4}\scriptsize{$\pm$1.5}~\gup{6.1} & \textbf{5.20}\scriptsize{$\pm$0.07}~\gdn{51.2\%} & \textbf{0.425}\scriptsize{$\pm$0.013}~\gup{77.1\%} \\
\methodhigh & 54.7\scriptsize{$\pm$0.3}~\gup{3.4} & 6.12\scriptsize{$\pm$0.22}~\gdn{42.6\%} & 0.380\scriptsize{$\pm$0.006}~\gup{58.3\%} \\
\midrule
\multicolumn{4}{l}{\textbf{Qwen3.5-4B}} \\
Vanilla ReAct & 27.5\scriptsize{$\pm$0.9} & 8.27\scriptsize{$\pm$0.11} & 0.161\scriptsize{$\pm$0.005} \\
OTC-PO & 28.1\scriptsize{$\pm$1.3}~\gup{0.6} & \textbf{4.14}\scriptsize{$\pm$0.29}~\gdn{49.9\%} & 0.223\scriptsize{$\pm$0.013}~\gup{38.5\%} \\
\groupdash
\methodlow & \textbf{31.9}\scriptsize{$\pm$0.9}~\gup{4.4} & 5.01\scriptsize{$\pm$0.16}~\gdn{39.4\%} & \textbf{0.239}\scriptsize{$\pm$0.008}~\gup{48.4\%} \\
\methodhigh & \textbf{31.9}\scriptsize{$\pm$0.9}~\gup{4.4} & 7.60\scriptsize{$\pm$0.41}~\gdn{8.1\%} & 0.198\scriptsize{$\pm$0.011}~\gup{23.0\%} \\
\bottomrule
\end{tabular}
\end{table}
\begin{table}[!t]
\centering
\footnotesize
\setlength{\tabcolsep}{2.9pt}
\caption{Results on four held-out multi-hop benchmarks (deterministic single run; question counts in
Section~\ref{sec:setup}). Best per column and size in bold.}
\label{tab:extended}
\begin{tabular}{l ccc ccc ccc ccc}
\toprule
& \multicolumn{3}{c}{2Wiki} & \multicolumn{3}{c}{HotpotQA} & \multicolumn{3}{c}{MuSiQue} & \multicolumn{3}{c}{Bamboogle} \\
\cmidrule(lr){2-4}\cmidrule(lr){5-7}\cmidrule(lr){8-10}\cmidrule(lr){11-13}
Method & SR$\uparrow$ & AC$\downarrow$ & HV$\uparrow$ & SR$\uparrow$ & AC$\downarrow$ & HV$\uparrow$ & SR$\uparrow$ & AC$\downarrow$ & HV$\uparrow$ & SR$\uparrow$ & AC$\downarrow$ & HV$\uparrow$ \\
\midrule
\multicolumn{13}{l}{\textbf{Qwen3.5-9B}} \\
Vanilla ReAct & \textbf{75.8} & 5.64 & 0.544 & \textbf{48.6} & 4.87 & 0.368 & \textbf{29.0} & 6.59 & 0.194 & \textbf{76.0} & 4.05 & 0.606 \\
OTC-PO & 70.6 & 2.24 & \textbf{0.627} & 48.0 & 1.59 & \textbf{0.442} & 25.6 & 2.23 & \textbf{0.227} & 68.8 & 1.38 & \textbf{0.640} \\
\groupdash
\methodlow    & 63.2 & \textbf{1.94} & 0.571 & 41.4 & \textbf{1.09} & 0.391 & 21.4 & \textbf{1.55} & 0.197 & 43.2 & \textbf{0.50} & 0.421 \\
\methodhigh   & 70.4 & 2.19 & \textbf{0.627} & 46.2 & 1.51 & 0.427 & 25.2 & 2.10 & 0.226 & 60.8 & 1.20 & 0.572 \\
\midrule
\multicolumn{13}{l}{\textbf{Qwen3.5-27B}} \\
Vanilla ReAct & 80.2 & 6.78 & 0.530 & \textbf{55.8} & 5.97 & 0.391 & 34.6 & 8.07 & 0.206 & 74.4 & 2.46 & 0.653 \\
OTC-PO & 66.8 & \textbf{2.11} & 0.598 & 38.8 & \textbf{0.98} & 0.369 & 29.2 & \textbf{2.78} & 0.251 & 45.6 & \textbf{0.42} & 0.446 \\
\groupdash
\methodlow    & \textbf{81.4} & 2.44 & \textbf{0.715} & 53.6 & 1.77 & \textbf{0.488} & \textbf{35.2} & 3.20 & \textbf{0.296} & 72.0 & 0.86 & \textbf{0.689} \\
\methodhigh   & 81.2 & 2.75 & 0.700 & 54.2 & 2.35 & 0.478 & 34.8 & 4.00 & 0.278 & \textbf{75.2} & 1.70 & 0.688 \\
\midrule
\multicolumn{13}{l}{\textbf{Qwen3.5-4B}} \\
Vanilla ReAct & 66.0 & 3.09 & 0.558 & 45.2 & 2.89 & 0.387 & 21.4 & 6.22 & 0.147 & 47.2 & 2.19 & 0.420 \\
OTC-PO & 52.0 & \textbf{1.51} & 0.481 & 30.6 & \textbf{0.77} & 0.294 & 15.6 & \textbf{1.11} & 0.147 & 33.6 & \textbf{0.47} & 0.328 \\
\groupdash
\methodlow    & 66.8 & 1.99 & 0.601 & 44.8 & 1.25 & \textbf{0.420} & 19.8 & 2.74 & \textbf{0.171} & 48.8 & 0.89 & 0.466 \\
\methodhigh   & \textbf{75.4} & 2.80 & \textbf{0.649} & \textbf{48.8} & 2.78 & \textbf{0.420} & \textbf{23.2} & 5.97 & 0.163 & \textbf{69.6} & 2.12 & \textbf{0.622} \\
\bottomrule
\end{tabular}
\end{table}

\textbf{Compared methods.} Vanilla ReAct is the untrained base model; the other three each represent one class of
Section~\ref{sec:related}. Budget-Prompt, our instance of the inference-time
class~\citep{bats2025,chen2026steering,fang2026budget}, only states a call budget in the prompt;
DEPO~\citep{chen2025depo} adds an efficiency reward to offline preference alignment; OTC-PO~\citep{wang2025otc} is the
online-RL case, closest to ours. All four share \method's base model, toolset and protocol; both trained baselines are
reimplemented as their papers describe. Budget-Prompt and DEPO are compared at 9B, OTC-PO at all three sizes.

\section{Main Results}
\label{sec:results}

\textbf{Main results on GAIA.}
Five of the six operating points in Table~\ref{tab:main} improve on their base model in success rate and call count
at once, raising success rate by 3.4 to 6.1 points ($p=0.01$ for the largest) and reducing calls by 8.1\% to 51.2\%; the sixth, the efficiency point at 9B, is
statistically indistinguishable from it in success rate ($p=0.89$) while calling 58.6\% less. On GAIA, compression therefore does not sacrifice
correctness. At 9B, Budget-Prompt and DEPO fall below the base model in success rate, and OTC-PO matches it; our two points both call less than OTC-PO by more than run-to-run variation. Token counts fall with the calls (Appendix~\ref{app:tokens}).

\textbf{Transfer to the held-out benchmarks.}
The efficiency point's compression carries over to the four multi-hop benchmarks (Table~\ref{tab:extended}), with a
size-dependent effect on success rate: at 27B neither point differs significantly from the base model; at 4B
neither falls significantly below it and the capability point rises significantly on three of four; at 9B both points
fall below it on all four. Most questions here are two-hop, leaving little redundancy to remove before necessary calls are affected, and 9B is compressed most (its efficiency point calls 23\% as often as the base model,
against 35\% at 27B and 48\% at 4B). OTC-PO over-compresses the same way at 27B and 4B, whereas ours keeps success rate essentially unchanged there at a somewhat higher call count.

\section{Analysis and Ablations}
\label{sec:analysis}

\subsection{Sources of the Call Reduction}
\label{sec:analysis:what}

\textbf{Over half the reduction comes from trajectories that exhaust the call cap.}
Trajectories that exhaust the $T_{\max}$ cap fall from 15.6\% of the base model's to 1.4\% and 3.4\% at the efficiency and capability points. Almost all of the base model's capped trajectories answer wrongly; calls spent in capped trajectories fall from 3.11 to 0.28 per question, 58.5\% of the efficiency point's reduction. Tail repetition loops (form (ii), Appendix~\ref{app:diagnosis}) often run to the cap; capped or not, they end 17.6\% of the base model's trajectories and account for 33\% of the reduction. Eq.~(\ref{eq:u}) charges such trajectories the most. By the diagnosis judge's per-call labels (Appendix~\ref{app:diagnosis}), of the calls the efficiency point removes, 58\% are form (ii), 20\% form (i) and 21\% necessary; form (iii), which belongs to no single call, stays under 1\%. The reduction is not abandonment---on the 37 questions that the base model and the efficiency point both answer correctly in every run, calls still fall by 57.5\%.

\textbf{Compression scales with task difficulty.}
Regressing per-question call counts on those of a separate base-model run (to avoid regression to the mean)
gives slopes of 0.91 for the base model itself and 0.22 and 0.26 for the efficiency and capability points (0.28 and
0.31 at 4B): the compressed call count still rises with the base model's, rather than every task being pulled
toward one common count. 

\subsection{Component Ablations}
\label{sec:analysis:effort}

The comparison with OTC-PO already contrasts its moving anchor, the group's shortest correct rollout, with our fixed
budget (Table~\ref{tab:main}, where one of our operating points has the higher HV at every size) and its collapsing run with
ours (Section~\ref{sec:analysis:collapse}); here we ablate the effort conditioning and the coverage term.

\begin{table}[!htb]
\centering
\small
\caption{Ablation of the effort conditioning (GAIA-165, 9B).}
\label{tab:effort}
\begin{tabular}{lccc l}
\toprule
Configuration & SR$\uparrow$ & AC$\downarrow$ & HV$\uparrow$ & Note \\
\midrule
Trained at $t=1$, low tag & 39.2 & 4.07 & 0.312 & same price at every tier \\
Trained at $t=1$, high tag & 40.6 & 4.91 & 0.306 & as above \\
Same weights, no tag at inference & 39.8 & 4.29 & 0.313 & the unconditioned tier from training \\
Untrained base + low tag & 37.0 & 7.01 & 0.240 & no training, prompt only \\
Untrained base + high tag & 40.6 & 7.77 & 0.248 & no training, prompt only \\
\midrule
\methodlow & 39.8 & \textbf{3.42} & 0.330 & efficiency operating point \\
\methodhigh & \textbf{43.8} & 4.44 & \textbf{0.341} & capability operating point \\

\bottomrule
\end{tabular}
\end{table}

\textbf{The operating points come from training rather than the prompt.}
Prompting the untrained base with the tags (Table~\ref{tab:effort}, rows 4--5) separates the call counts but changes
HV by less than 6\%: its low tag removes 1.25 calls at a loss of 3.2 points of success rate, whereas the trained
efficiency point removes 4.84 calls at a loss of 0.4 (Table~\ref{tab:main}). The tag specifies an effort level but not
which calls are redundant; training couples each tag to a per-call price, and the calls it removes are mostly those that
Eq.~(\ref{eq:u}) prices highest (Section~\ref{sec:analysis:what}). The unconditioned tier, trained at the intermediate
price $\gamma_\varnothing=1$, likewise lies between the two operating points in call count, consistent with each operating point
being set by the price under which it is trained.

\textbf{The capability point relies on the price factor.}
With $t=1$ (Table~\ref{tab:effort}, rows 1--2) all tiers are trained on one reward, so the residual separation in call
count stems from the tag alone. This separation no longer raises success rate: the high tag issues more calls than the full
method's capability point (4.91 against 4.44) at a lower success rate (40.6 against 43.8), and its margin over the low
tag shrinks from 4.0 points to 1.4. Only for $t>1$ does the high tier combine a lower per-call price with larger
partial credit for the evidence obtained, so that additional calls are rewarded for advancing the evidence chain
rather than for their number.

\textbf{The operating points differ on individual questions.}
The two operating points differ in call count on 112 of GAIA's 165 questions at 9B, and on 82 (73.2\%) the efficiency point
calls less (two-sided sign test, $p=9\times10^{-7}$); over the twelve combinations of four held-out benchmarks and
three model sizes, the direction holds on 79.2\%--100\% of the differing questions.

\label{sec:analysis:reward}

\textbf{The coverage term mainly affects incorrect trajectories.}
Removing the coverage term $(1-\mathrm{Cov})$ from Eq.~(\ref{eq:u}) while keeping partial credit leaves success
rate matching the full method (39.6 against 39.8) and raises the call count from 3.42 to 4.00. The increase is confined to
incorrect trajectories (3.97 to 4.90), while correct ones average 2.61 calls under both: they have mostly gathered
their evidence already, so the term adds little to their cost. 

\subsection{Call-Count Collapse in Training}
\label{sec:analysis:collapse}

\begin{figure}[!htb]
\centering
\includegraphics[width=\textwidth]{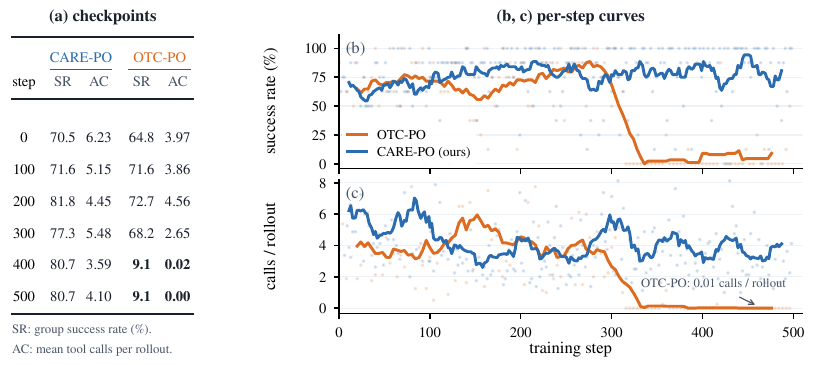}
\caption{Training dynamics at 27B for OTC-PO and our method, excluding the injected source trajectory; solid curves
are running means of the per-step values (dots). (a) The running means at six steps; bold values follow the collapse.
(b) Group success rate on the training distribution. (c) Mean tool calls per rollout.}
\label{fig:traindyn}
\end{figure}

\textbf{Unlike our method, OTC-PO at 27B collapses toward zero calls.}
After step 300 OTC-PO's call count falls to 0.01 per rollout, the degenerate solution of
Section~\ref{sec:method:anchor} (Figure~\ref{fig:traindyn}); at 9B and 4B it does not reach this solution in training,
while ours remains well above zero at all three sizes (Appendix~\ref{app:traindyn}). Tables~\ref{tab:main}
and~\ref{tab:extended} therefore evaluate 27B OTC-PO at an earlier checkpoint.

\section{Conclusion and Limitations}
\label{sec:conclusion}

We started from a step-level diagnosis: over 80\% of Qwen3.5's GAIA trajectories contain removable calls, at both sizes examined.
We treat call efficiency as a property controllable at inference and propose \method, which prices calls against a
reference budget anchored to each task's own difficulty and fixed before training, eases call-count collapse with
evidence pricing and a source-trajectory floor, and changes the price of a call by effort tier, so that one set of
weights yields an efficiency and a capability operating point. On GAIA, five of the six
operating points across three sizes improve on their base model in success rate and call count at once, and the sixth
matches it in success rate at 58.6\% fewer calls; over half the reduction comes from base-model trajectories that
exhaust the call cap, almost all of them failures. At all three sizes the group-mean call count remains well above zero
throughout training, whereas OTC-PO collapses toward issuing no call at 27B.

\textbf{Limitations and future work.} On the held-out multi-hop benchmarks, where most questions are
two-hop, compression readily encroaches on necessary calls: success rate does not fall significantly at 4B or 27B, but 9B,
compressed most, trades it for fewer calls at both points. Our reward presupposes automatically gradable goals and a
successful source trajectory for each training task; for lack of compute, training and evaluation cover only question
answering (GAIA and four short-answer multi-hop benchmarks), on one model family, Qwen3.5. Training and evaluating on
more complex tasks without a single success criterion, and estimating
reference budgets for tasks with no source trajectory, are natural next steps.

% ---- 以下三节均不计入正文页数（ICLR 2027 规定）--------------------------------
% The sections below do not count towards the main-text page limit (ICLR 2027 rules). The AI Use Statement is mandatory.
\section*{Ethics Statement}
Training tasks are drawn from the public MiroVerse dataset~\citep{miroverse2025}, and evaluation uses the public
benchmarks GAIA~\citep{mialon2023gaia}, 2Wiki~\citep{ho2020constructing}, HotpotQA~\citep{yang2018hotpotqa},
MuSiQue~\citep{trivedi2022musique} and Bamboogle~\citep{press2023compositionality}; no human subjects are involved.
Training and evaluation query the live web through third-party search and page-fetching APIs; the per-question records
we release contain no retrieved web content.

\section*{Reproducibility Statement}

Section~\ref{sec:method:bref} and Appendix~\ref{app:impl} describe how the training data, reference budgets and
evidence checklists are constructed; Section~\ref{sec:setup} gives the evaluation protocol and metrics, and
Appendix~\ref{app:impl} the hyperparameters, the toolset and the prompt specifications. The baselines that require
training (DEPO, OTC-PO) are reimplemented as their papers describe. The supplementary material contains the training
and rollout code with its prompts, a standalone implementation of the reward, the training configurations at the three
sizes, the final LoRA adapters at 9B and 27B, the list of training tasks, and the per-question evaluation records with
a script that recomputes Tables~\ref{tab:main}--\ref{tab:effort}.

\section*{AI Use Statement}

Large language models are used in three steps of this work, each described in the paper: the offline pass that
produces the reference budget, the evidence checklist and the gradable answer (Section~\ref{sec:method:bref}), the
coverage judgment during training (Section~\ref{sec:method:reward}), and the diagnosis of redundant calls
(Sections~\ref{sec:diagnosis} and~\ref{sec:analysis:what}). We also used LLM assistants for coding and data analysis,
literature search, and language editing, including drafting parts of the text. The authors reviewed all such output
and take full responsibility for the content of the paper.

\bibliography{refs}

\begin{thebibliography}{41}
\providecommand{\natexlab}[1]{#1}
\providecommand{\url}[1]{\texttt{#1}}
\expandafter\ifx\csname urlstyle\endcsname\relax
  \providecommand{\doi}[1]{doi: #1}\else
  \providecommand{\doi}{doi: \begingroup \urlstyle{rm}\Url}\fi

\bibitem[Aggarwal \& Welleck(2025)Aggarwal and Welleck]{l12025}
Pranjal Aggarwal and Sean Welleck.
\newblock {L1}: Controlling how long a reasoning model thinks with
  reinforcement learning.
\newblock In \emph{Conference on Language Modeling (COLM)}, 2025.
\newblock arXiv:2503.04697.

\bibitem[Bouzenia \& Pradel(2025)Bouzenia and
  Pradel]{bouzenia2025understanding}
Islem Bouzenia and Michael Pradel.
\newblock Understanding software engineering agents: A study of
  thought-action-result trajectories.
\newblock In \emph{40th IEEE/ACM International Conference on Automated Software
  Engineering (ASE)}, 2025.

\bibitem[Chen et~al.(2026{\natexlab{a}})Chen, Zhao, Xu, Zhao, Zhu, Zhang, Zhao,
  and Lu]{chen2025depo}
Sirui Chen, Mengshi Zhao, Lei Xu, Yuying Zhao, Beier Zhu, Hanwang Zhang,
  Shengjie Zhao, and Chaochao Lu.
\newblock {DEPO}: Dual-efficiency preference optimization for {LLM} agents.
\newblock In \emph{Proceedings of the AAAI Conference on Artificial
  Intelligence (AAAI)}, 2026{\natexlab{a}}.

\bibitem[Chen et~al.(2026{\natexlab{b}})Chen, Siu, Liu, Song, and
  Wang]{chen2026steering}
Yuqi Chen, Vincent Siu, Yang Liu, Dawn Song, and Chenguang Wang.
\newblock Tunable tool-call rates in {LLM} agents via representation steering.
\newblock \emph{arXiv preprint arXiv:2608.25198}, 2026{\natexlab{b}}.

\bibitem[Cuadron et~al.(2025)Cuadron, Li, Ma, Wang, Wang, Zhuang, Liu,
  Schroeder, Xia, Mao, Thumiger, Desai, Stoica, Klimovic, Neubig, and
  Gonzalez]{cuadron2025overthinking}
Alejandro Cuadron, Dacheng Li, Wenjie Ma, Xingyao Wang, Yichuan Wang, Siyuan
  Zhuang, Shu Liu, Luis~Gaspar Schroeder, Tian Xia, Huanzhi Mao, Nicholas
  Thumiger, Aditya Desai, Ion Stoica, Ana Klimovic, Graham Neubig, and
  Joseph~E. Gonzalez.
\newblock The danger of overthinking: Examining the reasoning-action dilemma in
  agentic tasks.
\newblock \emph{arXiv preprint arXiv:2502.08235}, 2025.

\bibitem[Deng et~al.(2026)Deng, Li, Gong, Ren, Thrampoulidis, and Li]{lld2025}
Wenlong Deng, Yushu Li, Boying Gong, Yi~Ren, Christos Thrampoulidis, and
  Xiaoxiao Li.
\newblock On group relative policy optimization collapse in agent search: The
  lazy likelihood-displacement.
\newblock In \emph{International Conference on Machine Learning (ICML)}, 2026.

\bibitem[Fang et~al.(2025)Fang, Ma, and Wang]{fang2025thinkless}
Gongfan Fang, Xinyin Ma, and Xinchao Wang.
\newblock {Thinkless}: {LLM} learns when to think.
\newblock In \emph{Advances in Neural Information Processing Systems
  (NeurIPS)}, 2025.

\bibitem[Fang et~al.(2026)Fang, Hu, Chang, Guo, Tao, Liu, Ruan, Huang, and
  Fang]{fang2026budget}
Zhengru Fang, Senkang~Forest Hu, Zhonghao Chang, Yu~Guo, Yihang Tao, Hongyao
  Liu, Mengzhe Ruan, Jun Huang, and Yuguang Fang.
\newblock Inference-time budget control for {LLM} search agents.
\newblock \emph{arXiv preprint arXiv:2605.05701}, 2026.

\bibitem[He et~al.(2025)He, Yuan, Li, Wang, and Chen]{thinkdial2025}
Qianyu He, Siyu Yuan, Xuefeng Li, Mingxuan Wang, and Jiangjie Chen.
\newblock {ThinkDial}: An open recipe for controlling reasoning effort in large
  language models.
\newblock \emph{arXiv preprint arXiv:2508.18773}, 2025.

\bibitem[He et~al.(2026)He, Sun, Cheng, Li, Ji, Lu, Huang, and Hu]{avspo2026}
Xixiang He, Qiyao Sun, Ao~Cheng, Xingming Li, Xuanyu Ji, Hailun Lu, Runke
  Huang, and Qingyong Hu.
\newblock Advantage collapse in group relative policy optimization: Diagnosis
  and mitigation.
\newblock In \emph{International Conference on Machine Learning (ICML)}, 2026.
\newblock arXiv:2605.21125.

\bibitem[Ho et~al.(2020)Ho, Nguyen, Sugawara, and Aizawa]{ho2020constructing}
Xanh Ho, Anh-Khoa~Duong Nguyen, Saku Sugawara, and Akiko Aizawa.
\newblock Constructing a multi-hop {QA} dataset for comprehensive evaluation of
  reasoning steps.
\newblock In \emph{Proceedings of the 28th International Conference on
  Computational Linguistics (COLING)}, 2020.

\bibitem[Hu et~al.(2022)Hu, Shen, Wallis, Allen-Zhu, Li, Wang, Wang, and
  Chen]{hu2022lora}
Edward~J. Hu, Yelong Shen, Phillip Wallis, Zeyuan Allen-Zhu, Yuanzhi Li, Shean
  Wang, Lu~Wang, and Weizhu Chen.
\newblock {LoRA}: Low-rank adaptation of large language models.
\newblock In \emph{International Conference on Learning Representations
  (ICLR)}, 2022.

\bibitem[Jin et~al.(2025)Jin, Zeng, Yue, Yoon, Arik, Wang, Zamani, and
  Han]{jin2025searchr1}
Bowen Jin, Hansi Zeng, Zhenrui Yue, Jinsung Yoon, Sercan Arik, Dong Wang, Hamed
  Zamani, and Jiawei Han.
\newblock {Search-R1}: Training {LLM}s to reason and leverage search engines
  with reinforcement learning.
\newblock In \emph{Conference on Language Modeling (COLM)}, 2025.

\bibitem[Li et~al.(2026)Li, Wang, Zhao, Chen, Li, Chen, Cao, Ye, Chai, and
  Yin]{li2026tepo}
Zeping Li, Hongru Wang, Yiwen Zhao, Guanhua Chen, Yixia Li, Keyang Chen, Yixin
  Cao, Guangnan Ye, Hongfeng Chai, and Zhenfei Yin.
\newblock Rethinking the role of entropy in optimizing tool-use behaviors for
  large language model agents.
\newblock In \emph{Annual Meeting of the Association for Computational
  Linguistics (ACL)}, pp.\  27955--27967, 2026.

\bibitem[Liu et~al.(2024{\natexlab{a}})Liu, Lin, Hewitt, Paranjape, Bevilacqua,
  Petroni, and Liang]{liu2024lost}
Nelson~F. Liu, Kevin Lin, John Hewitt, Ashwin Paranjape, Michele Bevilacqua,
  Fabio Petroni, and Percy Liang.
\newblock Lost in the middle: How language models use long contexts.
\newblock \emph{Transactions of the Association for Computational Linguistics},
  12:\penalty0 157--173, 2024{\natexlab{a}}.

\bibitem[Liu et~al.(2026{\natexlab{a}})Liu, Mao, Zhu, and
  Chua]{liu2026diagnosing}
Qi~Liu, Jiaxin Mao, Fengbin Zhu, and Tat-Seng Chua.
\newblock Diagnosing search behavior and failure modes in long-horizon search
  agents.
\newblock \emph{arXiv preprint arXiv:2608.01913}, 2026{\natexlab{a}}.

\bibitem[Liu et~al.(2026{\natexlab{b}})Liu, Wang, Miao, Hsu, Yan, Chen, Han,
  Xu, Chen, Jiang, Daruki, Liang, Wang, Pfister, and Lee]{bats2025}
Tengxiao Liu, Zifeng Wang, Jin Miao, I-Hung Hsu, Jun Yan, Jiefeng Chen, Rujun
  Han, Fangyuan Xu, Yanfei Chen, Ke~Jiang, Samira Daruki, Yi~Liang,
  William~Yang Wang, Tomas Pfister, and Chen-Yu Lee.
\newblock Budget-aware tool use enables effective agent scaling.
\newblock In \emph{Conference on Language Modeling (COLM)}, 2026{\natexlab{b}}.

\bibitem[Liu et~al.(2024{\natexlab{b}})Liu, Yu, Zhang, Xu, Lei, Lai, Gu, Ding,
  Men, Yang, Zhang, Deng, Zeng, Du, Zhang, Shen, Zhang, Su, Sun, Huang, Dong,
  and Tang]{liu2024agentbench}
Xiao Liu, Hao Yu, Hanchen Zhang, Yifan Xu, Xuanyu Lei, Hanyu Lai, Yu~Gu,
  Hangliang Ding, Kaiwen Men, Kejuan Yang, Shudan Zhang, Xiang Deng, Aohan
  Zeng, Zhengxiao Du, Chenhui Zhang, Sheng Shen, Tianjun Zhang, Yu~Su, Huan
  Sun, Minlie Huang, Yuxiao Dong, and Jie Tang.
\newblock {AgentBench}: Evaluating {LLM}s as agents.
\newblock In \emph{International Conference on Learning Representations
  (ICLR)}, 2024{\natexlab{b}}.

\bibitem[Ma et~al.(2026)Ma, Yang, Xu, Fang, Yu, and Chen]{ma2026trm}
Da~Ma, Ziyue Yang, Hongshen Xu, Haotian Fang, Kai Yu, and Lu~Chen.
\newblock Empowering {LLM} tool invocation with tool-call reward model.
\newblock In \emph{International Conference on Learning Representations
  (ICLR)}, 2026.

\bibitem[Ma et~al.(2025)Ma, Deng, Mao, Huang, Wang, Wu, Zhang, and
  Wang]{ma2025proofofuse}
Shengjie Ma, Chenlong Deng, Jiaxin Mao, Jiadeng Huang, Teng Wang, Junjie Wu,
  Changwang Zhang, and Jun Wang.
\newblock {Proof-of-Use}: Mitigating tool-call hacking in deep research agents.
\newblock \emph{arXiv preprint arXiv:2510.10931}, 2025.

\bibitem[Mialon et~al.(2024)Mialon, Fourrier, Swift, Wolf, LeCun, and
  Scialom]{mialon2023gaia}
Gr{\'e}goire Mialon, Cl{\'e}mentine Fourrier, Craig Swift, Thomas Wolf, Yann
  LeCun, and Thomas Scialom.
\newblock {GAIA}: A benchmark for general {AI} assistants.
\newblock In \emph{International Conference on Learning Representations
  (ICLR)}, 2024.

\bibitem[{MiroMind Team}(2025)]{miroverse2025}
{MiroMind Team}.
\newblock {MiroVerse}: A reproducible, full-trajectory, ever-growing deep
  research dataset.
\newblock \url{https://huggingface.co/datasets/miromind-ai/MiroVerse-v0.1},
  2025.
\newblock Hugging Face dataset, v0.1.

\bibitem[Press et~al.(2023)Press, Zhang, Min, Schmidt, Smith, and
  Lewis]{press2023compositionality}
Ofir Press, Muru Zhang, Sewon Min, Ludwig Schmidt, Noah~A. Smith, and Mike
  Lewis.
\newblock Measuring and narrowing the compositionality gap in language models.
\newblock In \emph{Findings of the Association for Computational Linguistics:
  EMNLP}, 2023.

\bibitem[Qian et~al.(2025)Qian, Acikgoz, Wang, Chen, Sil, Hakkani-T{\"u}r, Tur,
  and Ji]{qian2025smart}
Cheng Qian, Emre~Can Acikgoz, Hongru Wang, Xiusi Chen, Avirup Sil, Dilek
  Hakkani-T{\"u}r, Gokhan Tur, and Heng Ji.
\newblock {SMART}: Self-aware agent for tool overuse mitigation.
\newblock In \emph{Findings of the Association for Computational Linguistics:
  ACL}, 2025.

\bibitem[Qin et~al.(2024)Qin, Liang, Ye, Zhu, Yan, Lu, Lin, Cong, Tang, Qian,
  et~al.]{qin2024toolllm}
Yujia Qin, Shihao Liang, Yining Ye, Kunlun Zhu, Lan Yan, Yaxi Lu, Yankai Lin,
  Xin Cong, Xiangru Tang, Bill Qian, et~al.
\newblock {ToolLLM}: Facilitating large language models to master 16000+
  real-world {APIs}.
\newblock In \emph{International Conference on Learning Representations
  (ICLR)}, 2024.

\bibitem[{Qwen Team}(2025)]{qwen3}
{Qwen Team}.
\newblock {Qwen3} technical report.
\newblock \emph{arXiv preprint arXiv:2505.09388}, 2025.

\bibitem[{Qwen Team}(2026)]{qwen35}
{Qwen Team}.
\newblock {Qwen3.5}: Towards native multimodal agents.
\newblock \url{https://qwen.ai/blog?id=qwen3.5}, 2026.
\newblock Blog post, February 2026.

\bibitem[Schick et~al.(2023)Schick, Dwivedi-Yu, Dess{\`i}, Raileanu, Lomeli,
  Hambro, Zettlemoyer, Cancedda, and Scialom]{schick2023toolformer}
Timo Schick, Jane Dwivedi-Yu, Roberto Dess{\`i}, Roberta Raileanu, Maria
  Lomeli, Eric Hambro, Luke Zettlemoyer, Nicola Cancedda, and Thomas Scialom.
\newblock Toolformer: Language models can teach themselves to use tools.
\newblock In \emph{Advances in Neural Information Processing Systems
  (NeurIPS)}, 2023.

\bibitem[Shao et~al.(2024)Shao, Wang, Zhu, Xu, Song, Bi, Zhang, Zhang, Li, Wu,
  and Guo]{shao2024deepseekmath}
Zhihong Shao, Peiyi Wang, Qihao Zhu, Runxin Xu, Junxiao Song, Xiao Bi, Haowei
  Zhang, Mingchuan Zhang, Y.~K. Li, Y.~Wu, and Daya Guo.
\newblock {DeepSeekMath}: Pushing the limits of mathematical reasoning in open
  language models.
\newblock \emph{arXiv preprint arXiv:2402.03300}, 2024.

\bibitem[Shi et~al.(2026)Shi, Peng, Li, Wang, Wang, Du, and
  Zou]{shi2026overcalling}
Wei Shi, Ziheng Peng, Sihang Li, Xiting Wang, Xiang Wang, Mengnan Du, and
  Na~Zou.
\newblock To call or not to call: Diagnosing intrinsic over-calling bias in
  {LLM} agents.
\newblock \emph{arXiv preprint arXiv:2605.18882}, 2026.

\bibitem[Trivedi et~al.(2022)Trivedi, Balasubramanian, Khot, and
  Sabharwal]{trivedi2022musique}
Harsh Trivedi, Niranjan Balasubramanian, Tushar Khot, and Ashish Sabharwal.
\newblock {MuSiQue}: Multihop questions via single-hop question composition.
\newblock \emph{Transactions of the Association for Computational Linguistics},
  10:\penalty0 539--554, 2022.

\bibitem[Wang et~al.(2025)Wang, Qian, Zhong, Chen, Qiu, Huang, Jin, Wang, Wong,
  and Ji]{wang2025otc}
Hongru Wang, Cheng Qian, Wanjun Zhong, Xiusi Chen, Jiahao Qiu, Shijue Huang,
  Bowen Jin, Mengdi Wang, Kam-Fai Wong, and Heng Ji.
\newblock Acting less is reasoning more! {T}eaching model to act efficiently.
\newblock \emph{arXiv preprint arXiv:2504.14870}, 2025.

\bibitem[Wang et~al.(2026)Wang, Xie, Yang, Feng, Shen, Sun, Long, Jiao, Tan,
  Wang, Wei, and Gu]{webclipper2026}
Junjie Wang, Zequn Xie, Dan Yang, Jie Feng, Yue Shen, Duolin Sun, Meixiu Long,
  Yihan Jiao, Zhehao Tan, Jian Wang, Peng Wei, and Jinjie Gu.
\newblock {WebClipper}: Efficient evolution of web agents with graph-based
  trajectory pruning.
\newblock In \emph{Annual Meeting of the Association for Computational
  Linguistics (ACL)}, 2026.

\bibitem[Wei et~al.(2026)Wei, Pang, Liu, Shi, Deng, Xu, Duan, Wang, Sun, Shen,
  and Cheng]{wei2025rcp}
Zihao Wei, Liang Pang, Jiahao Liu, Wenjie Shi, Jingcheng Deng, Shicheng Xu,
  Zenghao Duan, Jingang Wang, Fei Sun, Huawei Shen, and Xueqi Cheng.
\newblock The evolution of thought: Tracking {LLM} overthinking via reasoning
  dynamics analysis.
\newblock In \emph{Annual Meeting of the Association for Computational
  Linguistics (ACL)}, 2026.

\bibitem[Wu et~al.(2026)Wu, Das, Amani, Nag, Lee, Gummadi, Ravichander, and
  Zafar]{wu2026tocall}
Qinyuan Wu, Soumi Das, Mahsa Amani, Arijit Nag, Seungeon Lee, Krishna~P.
  Gummadi, Abhilasha Ravichander, and Muhammad~Bilal Zafar.
\newblock To call or not to call: A framework to assess and optimize {LLM} tool
  calling.
\newblock \emph{arXiv preprint arXiv:2605.00737}, 2026.

\bibitem[Yan et~al.(2025)Yan, Li, Hu, Wang, Cui, Qu, Cheng, and
  Zhang]{yan2025luffy}
Jianhao Yan, Yafu Li, Zican Hu, Zhi Wang, Ganqu Cui, Xiaoye Qu, Yu~Cheng, and
  Yue Zhang.
\newblock Learning to reason under off-policy guidance.
\newblock In \emph{Advances in Neural Information Processing Systems
  (NeurIPS)}, 2025.

\bibitem[Yang et~al.(2018)Yang, Qi, Zhang, Bengio, Cohen, Salakhutdinov, and
  Manning]{yang2018hotpotqa}
Zhilin Yang, Peng Qi, Saizheng Zhang, Yoshua Bengio, William~W. Cohen, Ruslan
  Salakhutdinov, and Christopher~D. Manning.
\newblock {HotpotQA}: A dataset for diverse, explainable multi-hop question
  answering.
\newblock In \emph{Conference on Empirical Methods in Natural Language
  Processing (EMNLP)}, 2018.

\bibitem[Yao et~al.(2023)Yao, Zhao, Yu, Du, Shafran, Narasimhan, and
  Cao]{yao2023react}
Shunyu Yao, Jeffrey Zhao, Dian Yu, Nan Du, Izhak Shafran, Karthik Narasimhan,
  and Yuan Cao.
\newblock {ReAct}: Synergizing reasoning and acting in language models.
\newblock In \emph{International Conference on Learning Representations
  (ICLR)}, 2023.

\bibitem[Yu et~al.(2026)Yu, Tang, Liu, Zhang, Chu, and Han]{yu2026wedas}
Zixuan Yu, Zhenheng Tang, Tongliang Liu, Chengqi Zhang, Xiaowen Chu, and
  Bo~Han.
\newblock Rethinking deep research from the perspective of web content
  distribution matching.
\newblock \emph{arXiv preprint arXiv:2603.07241}, 2026.

\bibitem[Zhang et~al.(2025)Zhang, Lin, Hou, Feng, and Li]{zhang2025adaptthink}
Jiajie Zhang, Nianyi Lin, Lei Hou, Ling Feng, and Juanzi Li.
\newblock {AdaptThink}: Reasoning models can learn when to think.
\newblock In \emph{Conference on Empirical Methods in Natural Language
  Processing (EMNLP)}, 2025.

\bibitem[Zitzler \& Thiele(1999)Zitzler and Thiele]{zitzler1999hv}
Eckart Zitzler and Lothar Thiele.
\newblock Multiobjective evolutionary algorithms: a comparative case study and
  the strength {P}areto approach.
\newblock \emph{IEEE Transactions on Evolutionary Computation}, 3\penalty0
  (4):\penalty0 257--271, 1999.

\end{thebibliography}
\bibliographystyle{iclr2027_conference}

\appendix
\section{Diagnosing Redundant Calls: Procedure and Three Forms}
\label{app:diagnosis}

\begin{figure}[!t]
\centering
\includegraphics[width=\textwidth]{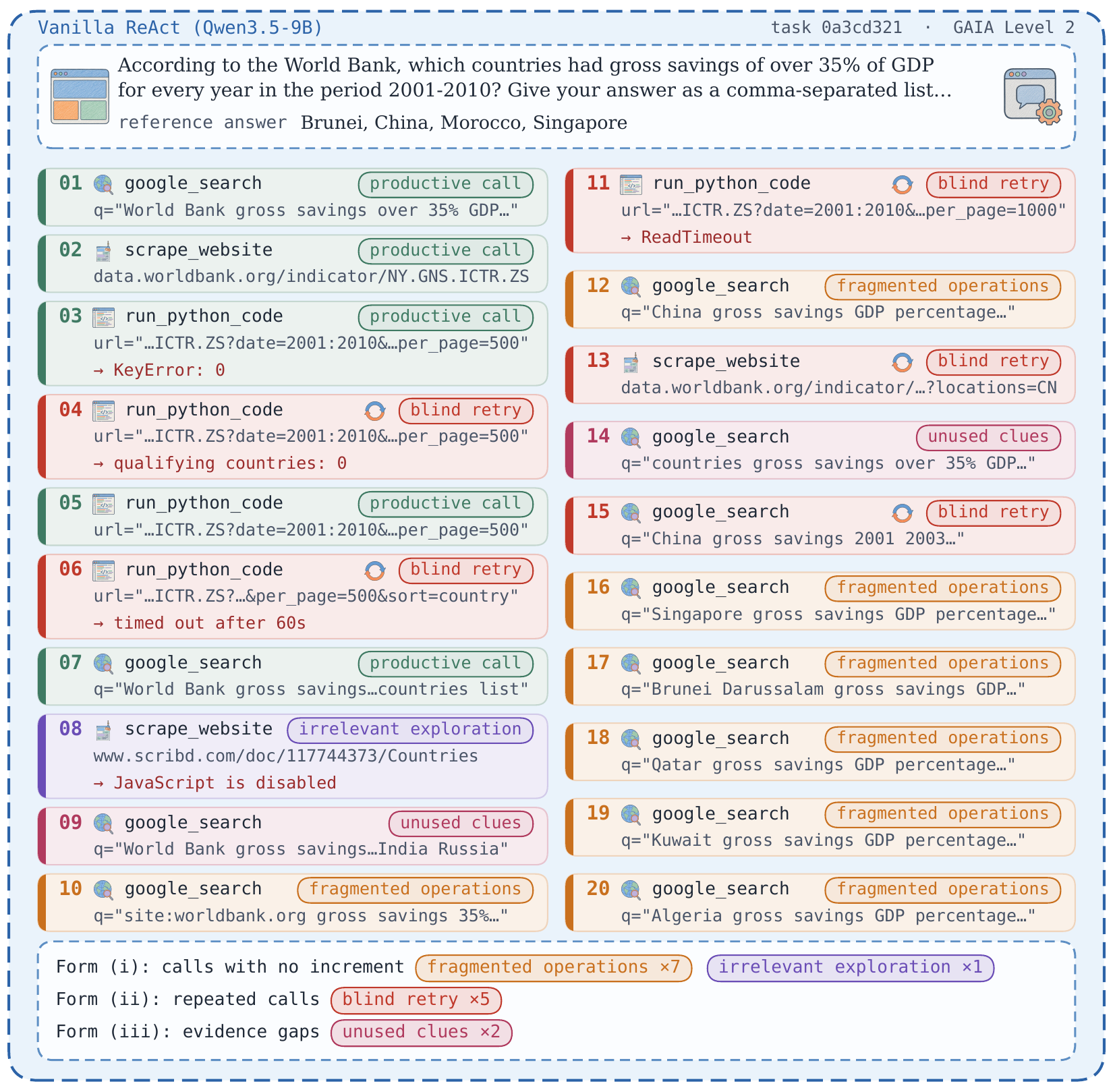}
\caption{A real trajectory of the 9B base model on GAIA task \texttt{0a3cd321} (Level 2), with its redundant calls
labelled step by step by the diagnosis judge. A call may match several forms; the label gives its principal one. The form (iii) cue is a
World Bank DataBank table with per-country values for each year of 2001--2010: its link appears in the search results
of steps 1, 7, 9, 10, 14 and 19 (snippets omitted), yet the table is never opened.}
\label{fig:diagcase}
\end{figure}

\textbf{Labelling criterion.} In the diagnosis of Section~\ref{sec:diagnosis}, a judge model examines each call in
turn: a call deletable outright or mergeable into a neighbouring call is labelled redundant. The judge is Qwen3.8-27B,
distinct from both diagnosed models.

\textbf{Form (i): calls with no increment.} The observation carries no new task-relevant information. Three
behaviours belong to this form: turning toward information unrelated to the goal, drawn by noisy feedback or a local cue
(\emph{irrelevant exploration}); continuing to confirm the same fact once the information already gathered suffices to
complete the task (\emph{over-verification}); and splitting into several calls what one more precise or combined call
would have obtained (\emph{fragmented operations}). The call count already measures them fully.

\textbf{Form (ii): repeated calls.} The call closely matches an earlier one in tool, arguments and return, and so
re-consumes an observation already obtained, the behaviour we label \emph{blind retry}. For tail loops and the consistency check below, a call counts
as a near-duplicate when it has the same tool name, a Jaccard similarity of at least 0.8 between the argument token
sets, and a Jaccard similarity of at least 0.9 between its return and some earlier return. In the training reward,
$S_{\mathrm{charged}}$ in Eq.~(\ref{eq:u}) counts only the unambiguous case: a call whose tool name and normalized
arguments exactly match those of an earlier call.

\textbf{Form (iii): evidence gaps.} A cue capable of advancing the task has appeared in the trajectory, yet the model
does not act on it and turns to tangential searches instead, so the evidence chain never closes. The gap itself belongs to no
single call; in step-by-step labelling, the label \emph{unused clues} marks the side searches launched after the cue appeared.
The coverage term of Eq.~(\ref{eq:u}) corresponds to this gap.

\textbf{Frequency and harm run in opposite orders.} Over the 183 trajectories that contain redundancy, where one
trajectory may match several forms, form (i) appears in 166 (90.7\%), form (ii) in 116 (63.4\%) and form (iii) in 76
(41.5\%); the error rates of those trajectories are 42.2\%, 68.1\% and 73.7\% respectively.

\textbf{Tail repetition loops are the most harmful case of redundancy.} A trajectory whose last three or more calls in
a row all meet the near-duplicate criterion above ends in a tail repetition loop; such a loop often runs on to the
call cap, leaving the trajectory unable to answer. The main finding of Section~\ref{sec:analysis:what}
rests on \emph{cap termination} (a trajectory exhausting the cap of $T_{\max}$ calls), which needs none of these
thresholds.

\textbf{Each of the three forms has a counterpart in prior work.} Diagnoses of deep-search
agents show that coarse queries draw in substantial irrelevant noise~\citep{yu2026wedas}, matching irrelevant
exploration; an empirical study of thought-action-result trajectories in software engineering agents records the
anti-pattern of repeating the same retrieval action without acting on its result~\citep{bouzenia2025understanding},
which on long-horizon tasks becomes a loop running to exhaustion~\citep{liu2024agentbench,webclipper2026}, matching
repeated calls and the tail repetition loop; the accumulation of redundant observations aggravates the difficulty
of using key information in long contexts~\citep{liu2024lost}, and evidence retrieved but not used correctly has been
identified as a failure in its own right~\citep{liu2026diagnosing}, matching evidence gaps.

\textbf{The three forms co-occur in one real trajectory.} Figure~\ref{fig:diagcase} labels a real trajectory step by
step: only 5 of its 20 calls advance the task, the rest fall into the three forms above, and the trajectory exhausts
the call cap without answering.

\begin{figure}[!t]
\centering
\includegraphics[width=\textwidth]{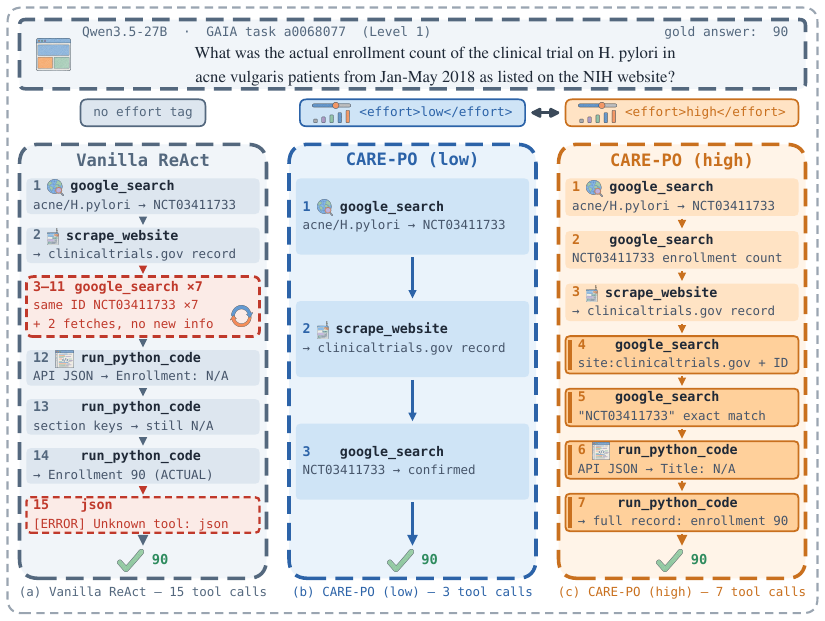}
\caption{Call trajectories of three policies on GAIA task \texttt{a0068077}, with arguments and observations
abbreviated. All three answer correctly; they differ in how many calls they need to reach the same
numeric answer. The highlighted
steps 4--7 in (c) are the capability point's extra checking calls.}
\label{fig:case}
\end{figure}

\textbf{Repeated calls are the largest source of the reduction.} For Section~\ref{sec:analysis:what}, the same judge
labels every call in all three GAIA runs of the 9B base model and both operating points. Given the question, the
reference answer and the whole trajectory, it writes a one-line rationale for each call and assigns it to necessary or
to one of the five behaviours above, which are then merged into forms. Per-category call counts are averaged over each
question's three runs before the reduction is split by category; resampling questions gives 95\% intervals of
49.8--65.8\%, 13.9--27.1\% and 15.5--28.0\% for the shares of form (ii), form (i) and necessary calls in the efficiency
point's reduction. Of all the base model's calls, 61.0\% are judged removable, close to the 68.0\% on the 221
separately collected trajectories of Section~\ref{sec:diagnosis}. The judge is also consistent with the deterministic
near-duplicate criterion above: among the 7964 calls it labelled, the criterion flags 1307, of which the judge marks
96.9\% as blind retries and 99.8\% as removable.

\section{Fine-Grained Results on GAIA}
\label{app:case}

Figure~\ref{fig:case} expands the task of Figure~\ref{fig:teaser}(b) (GAIA task \texttt{a0068077}, Level 1), placing side by side three real trajectories from the 27B evaluation logs.

The base model's first search already finds the trial and its second call fetches the page, yet it then takes two
kinds of detour: seven of calls 3 through 11 repeat searches for the same trial identifier, the \emph{blind retry} of
Appendix~\ref{app:diagnosis}, and calls 12 to 14 are three Python executions probing the same API response for the
right field name, a case of \emph{fragmented operations}.
\methodlow answers after 3 calls; it avoids both detours, adding only one confirming search after the fetch.
\methodhigh spends 7, all four extra calls used for checking: two further searches cross-check the identifier and two
Python executions re-verify the field.
On this task the two operating points use the same number of calls in all three independent runs (3 and 7), while the base
model uses 16, 12 and 15.

\textbf{Token cost decreases with the call count.}\label{app:tokens}
The reward prices tool calls, whereas inference cost is incurred in tokens. Table~\ref{tab:tokens} reports token
counts on the 9B GAIA traces, the setting analysed in Section~\ref{sec:analysis}. Relative to the base model, the
efficiency point generates 45\% fewer tokens, consumes 50\% fewer observation tokens and processes 67\% less context;
the capability point reduces these by 35\%, 38\% and 57\%, and OTC-PO by 26\%, 32\% and 49\%. The reduction in calls
is thus not offset by longer generation, and its effect accumulates because every observation is re-processed at each
subsequent turn. Each turn comprises one model call and at most one tool invocation; the number of these sequential
steps decreases by 52\% and 41\% at the two points.

\begin{table}[!htb]
\centering
\small
\caption{Per-question cost on GAIA-165 at 9B, averaged over the three runs of Table~\ref{tab:main}. Tokens are counted
with the Qwen3.5 tokenizer over the stored traces; \emph{context processed} is the total prompt length over the model calls
of a trajectory (system prompt, question and all preceding messages).}
\label{tab:tokens}
\begin{tabular}{lccccc}
\toprule
Method & AC & Turns & Generated & Observations & Context processed \\
\midrule
Vanilla ReAct & 8.26 & 9.1 & 1.91k & 13.1k & 104.4k \\
OTC-PO & 5.01 & 6.0 & 1.41k & 8.90k & 53.2k \\
\methodlow & 3.42 & 4.4 & 1.04k & 6.54k & 34.6k \\
\methodhigh & 4.44 & 5.4 & 1.24k & 8.11k & 45.3k \\
\bottomrule
\end{tabular}
\end{table}

\section{Training Dynamics}
\label{app:traindyn}

\begin{figure}[!htb]
\centering
\includegraphics[width=\textwidth]{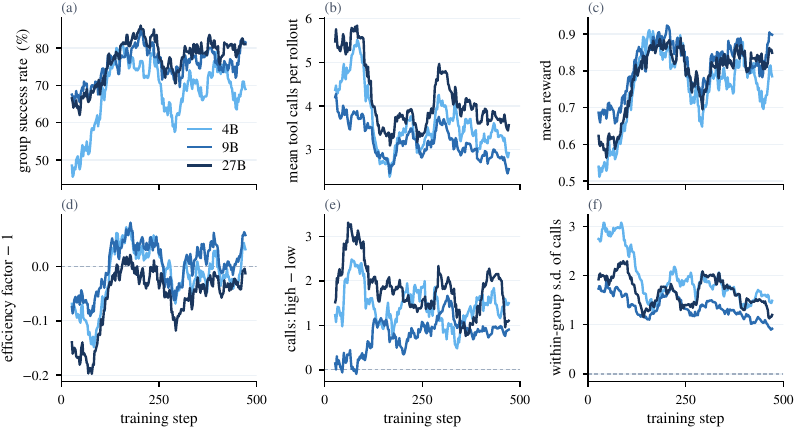}
\caption{Training dynamics of our method at three model sizes, excluding the injected source trajectory; curves are
centred running means over 25 optimization steps. (a) Group success rate on the training distribution. (b) Mean tool
calls per rollout. (c) Mean reward. (d) The efficiency factor minus one, i.e.\ $\lambda\tanh(\gamma_e(1-u))$ in
Eq.~(\ref{eq:reward}), which is zero (dashed line) at $u=1$, where cost is exactly one reference budget. (e) Difference
in mean call count between the high and low tiers. (f) Sample standard deviation of the call count within a group.}
\label{fig:traincurves}
\end{figure}

Figure~\ref{fig:traincurves} shows our method's training dynamics at all three sizes, including the composition of
the reward.

\textbf{Reward composition differs across sizes.}
The three sizes' mean rewards nearly coincide (Figure~\ref{fig:traincurves}(c)); they differ in the efficiency factor
(Figure~\ref{fig:traincurves}(d)). At 4B and 9B it rises from below one to around one; at 27B it also rises but remains
slightly below one for most of training, so 27B's comparable reward rests more on delivered value.

\textbf{The call count settles at a non-zero level at every size.}
Mean calls per rollout (Figure~\ref{fig:traincurves}(b)) fall early in training and then fluctuate well above zero; over the
last fifth of training, rollouts that issue no call at all make up only 2.22\%, 1.42\% and 0.28\% at 4B, 9B and 27B.

\textbf{Call counts rise with the reference budget.}
In the final fifth of training, delivering rollouts on tasks with reference budgets of 3, 4 and 5 use on average 2.31,
2.82 and 3.15 calls at 4B, 2.16, 2.77 and 2.91 at 9B, and 2.57, 3.51 and 3.54 at 27B, whereas no such ordering is
present in the first fifth. The policy thus acquires task-dependent call counts rather than converging to a single
count.

\textbf{The tiers separate early in training.}
The gap in call count between the tiers (Figure~\ref{fig:traincurves}(e)) is established within the first hundred steps and remains positive
at all three sizes thereafter.

\textbf{Call counts remain dispersed within groups.}
Over the policy's own rollouts, the within-group standard deviation of the call count (Figure~\ref{fig:traincurves}(f))
remains well above zero throughout training at every size, so the group-relative advantage retains a signal on calling.

\textbf{The stability advantage persists under longer training.}
When both methods are trained at 27B for 1000 steps with the same learning rate, OTC-PO again
ends up issuing no call at all, and its training success rate falls with it. Ours levels off at about three calls per
rollout with fewer than 1\% of rollouts issuing no call, and its training success rate does not decline.

\textbf{Every model in the result tables is evaluated at its final checkpoint.} The one exception is OTC-PO at 27B, whose final
checkpoint has already reached the degenerate solution (Figure~\ref{fig:traindyn}); we evaluate a checkpoint from
before the collapse, so that the comparison is not drawn against a collapsed policy.

\section{Implementation Details and Hyperparameters}
\label{app:impl}

\textbf{Training.} The optimizer is GRPO~\citep{shao2024deepseekmath} with no KL regularizer and token-level loss
normalization. Each group holds $G=8$ policy rollouts and the
injected source trajectory (Section~\ref{sec:method:anchor}), which enters the same clipped surrogate with the current
policy's log-probabilities as its old ones, so its importance ratio starts at 1. The reward uses $\lambda=0.5$, $t=1.5$
and $b_{\text{high}}=0.30$. LoRA~\citep{hu2022lora} of rank 8 is applied to the attention and MLP projections. Training
runs for 500 steps with a call cap of $T_{\max}=20$, at a learning rate of $6\times10^{-6}$ (4B, 9B) or $1\times10^{-5}$
(27B).

\textbf{Tools.} Training and evaluation share four tools: \texttt{google\_search} (web search),
\texttt{scrape\_website} (fetch the body text at a URL), \texttt{run\_python\_code} (execute code and return stdout) and \texttt{read\_file} (read a file attached to the task). The first two reach the live web through third-party APIs; the other two run in a local sandbox. All methods share one system prompt, which already instructs economical tool use. Tool signatures and return formats, the code and the verbatim prompts are in the supplementary material.

\textbf{Effort tag.} The low tier appends to the user turn a block headed \texttt{[Tool-call effort]} with the tag \texttt{<effort>low</effort>} and the sentence
``Prefer fewer tool calls; stop as soon as the evidence is sufficient''; the high tier appends
\texttt{<effort>high</effort>} and ``Allow extra tool calls when they are likely to improve confidence or recover
missing evidence''. The unconditioned tier adds neither, and evaluation uses the same text.

\textbf{Producing the reference budget and the evidence checklist.} The estimation prompt asks the judge to inspect the
source trajectory and return a minimal chain retaining only the essential steps and their evidence sources, together
with its length; entities drawn from the question or from common knowledge need no retrieval and are not counted in
the evidence checklist. The verification prompt presumes that the candidate is too low and checks
three conditions: that every entity a step requires traces back to the question or to an earlier observation, that the
evidence chain is coherent, and that the candidate is not below what is reachable in execution; if any check fails,
it returns a corrected value to the estimation step. An error in $B_{\mathrm{ref}}$ does not change the direction of
the pressure: the efficiency factor strictly decreases in $u$ and $u$ strictly increases in $S_{\mathrm{charged}}$, so
on a given task, among trajectories with the same outcome and the same coverage, the one with the smaller
$S_{\mathrm{charged}}$ scores higher for any $B_{\mathrm{ref}}$; the error changes only how much one call adds to
$u$, that is, the magnitude of the pressure.

\textbf{Training-task selection.} The three checks of Section~\ref{sec:method:bref} act on the output of the offline
pass: the evidence checklist must hold at least two entities that require retrieval
($|E(x)|\ge 2$), or coverage degenerates to a binary signal; the reference budget must be at least the length of that
checklist ($B_{\mathrm{ref}}\ge|E(x)|$), or a call that obtains one more piece of evidence costs more budget than the coverage it adds; and the
source trajectory must make at least twice as many calls as the reference budget ($S_{\mathrm{long}}\ge
2B_{\mathrm{ref}}$), or it leaves too little to compress. Of the candidates, 1825 pass all three checks; the
300-task training list is sampled from them with reference budgets of 3, 4 and 5 in about equal proportion. The listed
tasks have source-trajectory call counts of 7--12 (mean 10.5), reference budgets of 3--5 (mean 4.0) and checklists of length 2--3.

\textbf{Judging coverage.} For each entity on the checklist, a large model checks the trajectory's tool observations
for the entity or an equivalent rendering of it; the entity is covered if any observation contains it.

\textbf{Grading.} Evaluation decodes greedily. During training, answers are graded against the gradable answer produced by the offline pass of
Section~\ref{sec:method:bref}. At evaluation GAIA uses the official strict exact match (numeric equality, element-wise list
comparison, normalized string equality) and the multi-hop benchmarks use cover-EM, which counts an answer correct when the normalized gold string occurs in the
normalized prediction. The answer is taken from the closing
JSON object the prompt requires; where a trajectory does not end in one we try a \texttt{\textbackslash boxed\{\}}
expression and then a fallback extractor, applied identically to every method.

\textbf{Significance tests.} On GAIA we use a paired bootstrap over the 165 questions: 20000 resamples, each taking
the mean per-question difference in grades between two methods, with a two-sided $p$; on the single-run
multi-hop benchmarks we use an exact McNemar test on the paired per-question outcomes. 

\textbf{The exchange rate implied by HV.} $\mathrm{HV}=\mathrm{SR}\cdot(1-\mathrm{AC}/T_{\max})$ carries an implied
exchange rate. Differentiating at the 9B efficiency point (SR 39.8, AC 3.42) gives
$\partial\mathrm{HV}/\partial\mathrm{SR}=0.0083$ per point and $\partial\mathrm{HV}/\partial\mathrm{AC}=-0.0199$ per
call: one call saved is worth about 2.4 points of success rate. The rate is set by $T_{\max}$, and a larger cap makes
a call cheaper. This conclusion is robust to the rate: when HV is recomputed with $T_{\max}=30$, 50 or 100, the
highest HV at each size is still attained by one of our two operating points. We use HV to report results only, not to select checkpoints (Appendix~\ref{app:traindyn} describes how
checkpoints are chosen).

\end{document}